\documentclass{article}

\usepackage{arxiv}
\usepackage[authoryear]{natbib}

\usepackage{fancyhdr,graphicx,amsmath,amssymb}
\usepackage{algorithm, algorithmic}
\usepackage{url}
\usepackage{hyperref}
\include{pythonlisting}
\usepackage{caption}
\usepackage{subcaption}

\usepackage{setspace} 
\usepackage{lineno}
\usepackage[utf8]{inputenc} 
\usepackage[T1]{fontenc}    
\usepackage{hyperref}       
\usepackage{url}            
\usepackage{booktabs}       
\usepackage{amsfonts}       
\usepackage{nicefrac}       
\usepackage{microtype}      
\usepackage{lipsum}		
\usepackage{graphicx}
\usepackage{natbib}
\usepackage{doi}

\title{Explainable AI Integrated Feature Selection for Landslide Susceptibility Mapping using TreeSHAP}

\author{{Muhammad Sakib Khan Inan}\thanks{Corresponding Author} \\
	Department of Computer Science and Engineering\\
	East Delta University\\
	Chittagong, Bangladesh \\
	\texttt{sakib.khaninan@gmail.com} \\
	\And
	{Istiakur Rahman} \\
	Department of Civil and Environmental Engineering\\
	Islamic University of Technology (IUT) \\
	Gazipur, Dhaka, Bangladesh \\
	\texttt{istiakur92@iut-dhaka.edu} \\
}

\renewcommand{\shorttitle}{\textit{Landslide Susceptibility Mapping Explainable Artificial Intelligence}}

\begin{document}
\maketitle

\begin{abstract}
Landslides have been a regular occurrence and an alarming threat to human life and property in the era of anthropogenic global warming. An early prediction of landslide susceptibility using a data-driven approach is a demand of time. In this study, we explored the eloquent features that best describe landslide susceptibility with state-of-the-art machine learning methods. In our study, we employed state-of-the-art machine learning algorithms including XgBoost, LR, KNN, SVM, and Adaboost for landslide susceptibility prediction. To find the best hyperparameters of each individual classifier for optimized performance, we have incorporated the Grid Search method, with 10 Fold Cross-Validation. In this context, the optimized version of XgBoost outperformed all other classifiers with a Cross-validation Weighted F1 score of 94.62\%. Followed by this empirical evidence, we explored the XgBoost classifier by incorporating TreeSHAP, a game-theory-based statistical algorithm used to explain Machine Learning models, to identify eloquent features such as SLOPE, ELEVATION, TWI that complement the performance of the XGBoost classifier mostly and features such as LANDUSE, NDVI, SPI which has less effect on models performance. According to the TreeSHAP explanation of features, we selected the 9 most significant landslide causal factors out of 15. Evidently, an optimized version of XgBoost along with feature reduction by 40\% has outperformed all other classifiers in terms of popular evaluation metrics with a Cross-Validation Weighted F1 score of 95.01\% on the training and AUC score of 97\%.
\end{abstract}


\section{Introduction}
\label{sec:intro}
Landslides are subject to great concern for Geological and Environmental researchers from all over the world due to their irreparable and execrable impact on the environment, society, and economy. A landslide is a natural calamity that is characterized by the movement of a mass of rock, debris, or earth down a slope. Several environmental factors like heavy rainfall, geographical factor like location, land near volcanoes, etc contribute to the occurrence of landslides. Especially, in some areas, Rainfall, slope, Land Use Land Cover Change, Elevation, etc are one of the most influencing factors for Landslide \citep{Mindje2020}. \\

Hilly and Coastal areas all over the globe are mostly vulnerable to frequent and devastating landslides. More than a thousand people are killed by landslides every year around the globe, including an average of 25 - 50 deaths all alone in the United States. \citep{us-news-landslides}. According to the study of \citet{neegar-survey}, considering the time period of 2000-2018, the yearly average number of landslides in Bangladesh is 19, with a 4\% rate of increase per year, which ultimately results in 38 fatalities and 54 injuries on average. Landslides often damage road networks in hilly areas causing great direct or indirect consequential economic losses due to hindrance in communication with the hilly parts of a country \citep{landslide-road-net}. \\

Landslides are a great threat to the socio-economic conditions of a country \citep{Perera2018}. The impact of a landslide may be extended to the destruction of important infrastructure, cultivable land, and natural resources. It may lead to the blockage of rivers and intensify the risk of floods \citep{FAO}. The study Landslide Susceptibility is sensitive and arduous due to the presence of uncorrelated or non-linearly correlated environmental factors responsible for Landslides \citep{Huang2020}. In this context, an extensive statistical data-driven analysis could help us to detect useful hidden and confounding patterns for the identification of landslides to support effective measures to prevent this disaster. \\

In recent years, the global issue of Landslide, an alarming threat to mankind, has drawn attention of the Artificial Intelligence researchers. Considering the exigency of an early and automated prediction of Landslides, along with Geologists and Environmental scientists, AI (Artificial Intelligence)  researchers from all over the world have devoted themselves to the extensive study of Landslide Susceptibility Mapping using Artificial Intelligence based methods. As a result, over the years a decent number of state of art studies have been conducted for Landslide Susceptibility Mapping with Machine Learning, Deep Learning, and Artificial Neural Networks. The immense potential of Machine Learning algorithms can be utilized to automate and improve the efficiency of the analysis and prediction of Landslide Susceptibility \citep{Sahin-AI-tool}. \\

Despite a large number of outstanding researches have been conducted on Landslide Susceptibility using Machine Learning and Deep Learning methods, no study has solved the issue of explainability of these state of art  Artificial Intelligence algorithms. As the structure of these state of art algorithms are very complex from a mathematical point of view, it often engenders a BlackBox problem which may sometimes lead to the inefficiency of these framework in the future. And also, no previous studies have dedicatedly investigated the possibility of successful prediction of Landslides using less number of geological and topological factors with the help of machine learning-based automated systems. In this concern, in this study, we have employed an Explainable Artificial Intelligence-based framework for predicting the Susceptibility of Landslides. The key contributions of this study are as follows:

\begin{itemize}
    \item Identification of most important geological, topological, and hydrological factors that best corroborate the performance of automated machine learning model for the assessment and early prediction of Landslide Susceptible Area.
    \item Extensive performance analysis of machine learning classifiers to identify the most suitable Machine Learning algorithm for landslide susceptibility mapping.
    \item Proposal of an optimized integrated ensemble-based machine learning solution which needs less number of landslide causal features to efficiently classify Landslides by reducing costs and time consumed in the primary stage Landslide related studies.
\end{itemize}

The rest of the paper is structured as follows: section \ref{sec:related-works} highlights the related works by analyzing the existing state-of-the-art studies, section \ref{sec:method} describes algorithms or methods incorporated towards the development of our proposed automated machine learning model, section \ref{sec:dataset} analyzes and represents the dataset and features, section \ref{sec:results} illustrates the experimental results of statistical and machine learning based analysis, section \ref{sec:discussion} discuss and presents validation for the experimental results and highlights the significant research findings followed by a conclusion in section \ref{sec:conclusion}.

\section{Related Works}
\label{sec:related-works}

Machine Learning methods including Naïve Bayes (NB), Multilayer Perceptron (MLP), Kernel Logistic Regression (KLR), and J48-bagging employed for Landslide Susceptibility Mapping considering the area of Youfang district, China in the study of \citet{Hong2019} and it is observed from the study that MLP(Multilayer Perceptron) outperformed all other classifiers and proved to be an efficient tool for landslide study of this area. \citet{mehebub} proposed a hybrid neural network classifier integrating Multilayer Perceptron (MLP) and Bagging for an efficient mapping susceptibility of rainfall-induced landslides to support the identification of vulnerable areas for disaster prevention and management. Several other studies have also found that Multilayer Perceptron(MLP) or a hybrid combination of Multilayer Perceptron(MLP) and Particle Swarm Optimization is a compelling classifier in the study of Landslides \citep{Pham2017, Li2019}. \\

Deep Learning algorithms excelled in the study of Landslide Susceptibility prediction and analysis due to their deep architectures supporting robust in-built feature extraction strategies and a great capacity to tackle confounding and sensitive factors. \citet{Huang2020} found in their study that a Fully Connected Sparse Autoencoder-based model outperformed the traditional models for extracting optimal non-linear features from environmental factors. Moreover, a spatially explicit deep neural network was proposed by \citep{Dong}, where for feature selection Relief-F method was integrated to quantify the utility of the conditioning factors. Deep Neural Networks were found to be comparatively outperforming the conventional machine learning classifiers in the domain of Landslide Susceptibility in several other states of art studies \citep{BUI2020, Zhu2020}. \\

Though Deep Learning based methods can efficiently predict Landslides, the architectures of these algorithms are complex in structure and computationally expensive. Even for training purposes, to build an effective model, deep neural networks need a large amount of training data. In this context, to find an efficient but less computationally expensive framework for Landslide study, researchers have studied the potential of Ensemble-based, Probabilistic, and Hybrid Machine Learning classifiers for Landslide Susceptibility Prediction. Bagging-based Reduced Error Pruning Trees (BREPT), a novel hybrid machine learning classifier, was proposed by \citet{PHAM2019} with notable performance in Landslide Susceptibility. Similarly, another tree-based hybrid classifier was proposed by \citet{Saro2019} with impressive AUC scores. Also, the Classification and Regression Tree (CART) algorithm outperformed Multilayer Perceptron-based classifiers in some recent studies\citep{HUANG-ML2020}. \\

Some state art studies outlined the extreme potential of ensemble-based or gradient boosting-based machine learning classifiers through comparative analysis of ensemble machine learning methods and gradient boosting algorithms with several other methods \citep{Fang, Sahin}. Tree-based ensemble algorithms showed prominent performance; Random Forest led with impressive results in the study of \citet{MERGHADI2020}. Hybrid Random Forest-based models with GeoDetector and RFE for factor optimization were developed by \citep{ZHOU2021}. Extreme Boosting algorithms tend to outperform in comparison with linear classifiers by robust feature extraction and the ability to deal with outliers \citep{Rabby-Dataset}. However, Support Vector Machine and Logistic Regression exhibited significant performance in Landslide Susceptibility prediction and also outperformed ANN(Artificial Neural Network) in some studies \citep{Chen2018, Kalantar}. A novel machine learning method with the integration of unsupervised machine learning method K- Means Clustering and supervised Decision Tree based classifier was proposed by \citet{GUO2021}.

\section{Methods}
\label{sec:method}
\begin{figure*}[h!]
	\centering
	\includegraphics[scale=0.5]{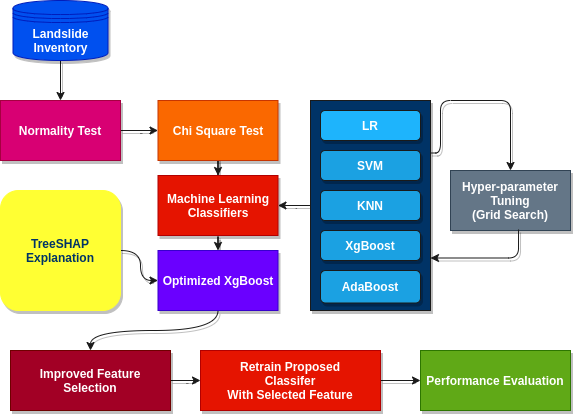}
	\caption{Our Research Framework}
	\label{framework}
\end{figure*}

In this section, an illustration of the methods that are implied in our study has been delineated in a detailed manner. The research framework of our study is graphically represented in Figure \ref{framework}.

\subsection{Machine Learning Classifiers}
In our study, we have employed 5 machine learning algorithms for experimental purposes and optimized their performance considering popular evaluation metrics. 
\subsubsection{Support Vector Machine}
Support Vector Machine (SVM) is a machine learning classifier that works in a supervised manner. It delineates a boundary to disparate data points by analyzing the data points from the training set based on differences in data distribution across individual classes. This boundary is called decision boundary which helps to classify between classes like Landslide Susceptible or Non-Landslide Susceptible. The decision boundary of SVM is a hyperplane in an N-dimensional space which evidently classifies the data points of different classes. Data points that are closer to the hyperplane and evince a significant impact on the hyperplane’s direction and orientation are called support vectors. The performance of the SVM is optimized through support vectors. The weights associated with these support vectors corroborate the classifier to shape to orientations of the hyperplane. SVM is an empirically popular classifier that has exhibited noteworthy performance in the domain of Landslide Susceptibility Mapping \citep{SVM-Review}.

SVM can be further divided into two major categories, Linear SVM and Non-Linear SVM.  The category of SVM is decided based on the kernel function it uses to draw the hyperplane. In Non-Linear SVM, the radial basis function is computed as, for Kernel, K,
\begin{equation}\label{svm}
 K(x, y) = e^{-\frac{\left | x  - y \right |}{2\sigma ^{2}}}
\end{equation}
            
Where $x$ and $y$ are data points and $ \sigma $ is a free parameter that controls the degree of generalization. For better optimization, the generalization parameter and kernel function is tweaked to find the combination that best supports the classification of Landslide Susceptibility.
\subsubsection{Logistic Regression}
Logistic Regression is a linear machine learning classifier, well suitable for handling categorical and numerical feature variables and capable of robustly predicting binary outcomes. Logistic Regression implies a log odds ratio which is an iterative maximum likelihood method to predict whether a given set of features belongs to a certain class. Logistic Regression follows the logistic function or sigmoid function to draw a decision boundary between the two classes, Landslide Susceptible area, and Non-Landslide Susceptible area. For a univariate logistic regression, To state mathematically,

\begin{equation}
y = \frac{e^{(w_0 + w_1*x)}}{1 + e^{(w_0 + w_1*x)}}
\end{equation}

Where $y$ is the target variable, $w_0$ indicates bias, and, $w_1$ indicates the coefficient or weights of feature variable $x$. Logistic Regression is widely used in the domain of Landslide Susceptibility Mapping due to its simple structure and effective performance for binary classification problems \citep{logistic}.
\subsubsection{K Nearest Neighbours}
K Nearest Neighbours (KNN) is a non-parametric machine learning algorithm. It is also called a lazy learning algorithm, much more suitable for smaller datasets \citep{knn_clf}. To classify data instances into certain categories, KNN employs a neighborhood similarity analyzing method incorporating distance metrics like Manhattan Distance, Euclidean Distance, etc, to successfully create discriminating supervised clusters of individual class labels like Landslide Susceptible or Non-Landslide Susceptible. For the final classification of Landslide Susceptibility, it would follow majority voting criteria, considering the class labels of a certain number of nearest neighbors based on the results of distance metrics analysis. The generalized formula for computing distances for KNN is;
\begin{equation}
    Distance\left(X,Y\right)=\left(\sum _{i=1}^{n}|x_{i}-y_{i}|^{p}\right)^{\frac {1}{p}}
\end{equation}
Here, $x$ and $y$ are data points according to the feature variable and target variable, and $p$ controls which distance metrics to be used. For Manhattan Distance, $p$ is set to 1 and for Euclidean Distance $p$ is set to 2.
\subsubsection{AdaBoost}
Adaptive Boosting (AdaBoost) is a tree-based machine learning classifier, which aggregates multiple weak learners (decision trees) in an ensemble manner to build a robust classifier \citep{adaboost}. A single decision tree or weak learner in Adaboost is called a stump which has a maximum depth of 1 with one root and two leaves. The algorithm assigns more weight to difficult-to-classify instances and less to those which are well-classified. Thus, it creates a forest of stumps to efficiently classify a Landslide Susceptible and Non-Landslide Susceptible area. The amount of say for each stump is determined based on the classification error which is basically the sum of weights for incorrectly classified samples \citep{adaboost-explain}.
The number of stumps to use for classification has a significant impact on classification performance. It is an extremely fast classifier compared to other tree-based classifiers which makes it a strong candidate for landslide susceptibility prediction. 

\subsubsection{Extreme Gradient Boosting}
Extreme Gradient Boosting (XgBoost) algorithm is a tree-based ensemble learning classifier that solves the overfitting issue, commonly found in other decision tree-based classifiers with its improved gradient boosting strategy with built-in regularization and impressive gains in speed \citep{xgboost}. The improved regularization strategy makes this algorithm so robust that it has evinced notable performance for solving problems that include a large amount of unstructured data (text, images, etc) or dataset containing outliers and features that are sensitive in nature \citep{inan}. In comparison to the previous general Gradient Boosting (GB), a number of new design features such as robustness in handling missing values, approximate and sparsity-aware split-finding algorithm, parallel computing, cache-aware access, block compression, and sharding have made XgBoost an effective choice for complex and sensitive classification problems. 

The gradient descent method is employed to build every decision tree uniquely, followed by beginning with a certain  threshold and modifying the weights in an iterative manner by minimizing residuals in every single iteration. So, the trees built after every iteration remains unique as the error or mistakes done by the previous tree is minimized or regularized in the next tree to build. 

Mathematically, residuals are calculated to tackle the problem of unique trees. Residuals are errors between observed and predicted values. Each tree starts with a single leaf and all of the residuals go to the leaf.

\begin{equation}
\label{cover}
	\text{Cover,C} = \sum_{i=1}^{N}[\text{P(lf)}_i(1-\text{P(lf)}_i)]
\end{equation} 
Here, $P(lf)$ denotes previously predicted probability for ith leaf. 
Similarity Score, $S$, is calculated for each new leaf.
\begin{equation}
\label{similarity}
	S = \frac{\sum_{i=1}^{N}\text{R}}{\text{C}+\lambda}
\end{equation}
Here, in the equation of (\ref{similarity}), $\lambda$ is the regularization parameter that controls the pruning of trees, and $R$ and $C$ denote residuals and cover accordingly.

The Gain of the trees is computed by calculating the similarity scores of left, right, and root nodes. It tells us where to split the data.
\begin{equation}
\label{final-gain}
	\text{G} = 	\text{L}_{\text{S}} + \text{R}_{\text{S}} - \text{N}_{\text{S}}
\end{equation}

Here, $G$ stands for Gain, $L_S$, $R_S$, $N_S$ denote accordingly, the similarity score of Left, Right, and Root Node. 


\subsection{TreeSHAP Explanation}
TreeSHAP is an extension of  the SHAP (SHapley Additive exPlanations) method which elucidates the prediction or output of Machine Learning algorithms by computing Shapley values for a given data instance that delineates what is the sum of contributions from its individual feature variables \citep{SHAP, TreeSHAP}. It is a game theoretic approach instigated as a fast and model-specific alternative to KernelSHAP for decision tree-based algorithms. Shapley values, a coalitional game theory technique, illustrates how to allocate the prediction of individual instances among the characteristics in a fair manner. Shapley values are computed as \citep{Shapely-values};

\begin{equation}
    \phi_i = \sum_{S\subseteq {N - i}}^{}{\frac{|S|!(M - |S| - 1)!}{M!}[f_x(S\bigcup {i})-f_x(S)]}
\end{equation}

Here, $f(x)$ denotes the prediction function of the Machine Learning classifier, and $M$ stands for the total number of features.$S$ represent any subset of features that doesn’t include the $i-th$ feature and $|S|$ is the size of that subset.

In our study, we have incorporated the TreeSHAP method for analyzing the predictions of the Extreme Gradient Boosting algorithm to solve the black-box issue of machine learning algorithms in a motivation to build an Explainable Artificial Intelligence based solution along with identifying which of the features corroborates mostly an effective classification of Landslide Susceptibility.

\section{Dataset Description}
\label{sec:dataset}
In our study, we have experimented with our research methods on a benchmark dataset of Landslide Susceptibility Mapping considering three Upazilas of Rangamati Hill District, Bangladesh which was prepared by
\citet{Rabby-Dataset}. This district is extremely susceptible to landslides due to its geographical location and several natural factors. The natural climate factors like heavy rainfall and geographical factor-like the highest average slope gradients have made this district prone to landslides and a convincing candidate in terms of the study area for landslide research.
The following dataset contains 196 data instances of each of the two target classes, Landslide Susceptible and Non-Landslide Susceptible. 15 important Geological, Topological, and Hydrological factors for Landslide Susceptibility Mapping are presented in this dataset. In our study, we have included these 15 landslide causal factors as feature variables for predicting Landslide Susceptibility. The following feature variables are; 'PROFILE' (Profile Curvature), 'PLAN' (Plan Curvature), 'CHANGE','LANDUSE', 'ELEVATION', 'SLOPE', 'ASPECT', 'TWI', 'SPI', 'DRAINAGE' (Distance from Drainage Network), 'NDVI' (Normalized Vegetation Index), 'RAINFALL', 'FAULTLINES' (Distance to Fault lines), 'ROAD' (Distance to Road Network), 'GEOLOGY'. Table~\ref{tab:gis-methods} delineates the methods utilized to compute these landslide causal factors by incorporating remote sensing technologies \citep{Abedin_Features}. A brief description of the features which have been analyzed and utilized in this study is given below:
\begin{table*}[h!]
\centering
\begin{tabular}{|c|r|}
\hline
\textbf{Landslide Casual Factor} & \textbf{Methods Used to Determine the Factor}                                         \\ \hline
PROFILE CURVATURE       & ArcGIS Curvature Function                                                    \\ \hline
PLAN CURVATURE          & ArcGIS Curvature Function                                                    \\ \hline
LAND COVER CHANGE       & Anderson scheme Level-I method (Satelite Images)                             \\ \hline
LANDUSE CHANGE          & Anderson scheme Level-I method (Satelite Images)                             \\ \hline
ELEVATION & \begin{tabular}[c]{@{}r@{}}Advanced Spaceborne Thermal Emission and \\ Reflection Radiometer (ASTER) Global Digital\\ Elevation Map (GDEM)\end{tabular} \\ \hline
NDVI                    & LandSat 8 level 2 imagery                                                    \\ \hline
RAINFALL                & \begin{tabular}[c]{@{}r@{}}Kriging Interpolation\end{tabular}    \\ \hline
FAULTLINES              & \begin{tabular}[c]{@{}r@{}}Euclidean\\ Distance tool in ArcGIS\end{tabular}  \\ \hline
SLOPE                   & Slope tool in ArcGIS                                                         \\ \hline
ASPECT                  & Aspect tool in ArcGIS                                                        \\ \hline
TWI                     & ASTER GDEM                                                                   \\ \hline
SPI                     & ASTER GDEM                                                                   \\ \hline
DRAINAGE                & \begin{tabular}[c]{@{}r@{}}Euclidean\\ Distance tool  in ArcGIS\end{tabular} \\ \hline
ROAD                    & \begin{tabular}[c]{@{}r@{}}Euclidean\\ Distance tool  in ArcGIS\end{tabular} \\ \hline
GEOLOGY                 & Geological Survey                                                            \\ \hline
\end{tabular}
\caption{Methods To Compute Landslide Causal Factors}
\label{tab:gis-methods}
\end{table*}

\subsection{PROFILE (Profile Curvature)}
\begin{figure}[h!]
	\centering
	\includegraphics[scale = 0.6]{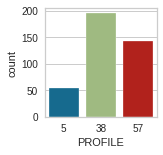}
	
	\caption{Count Distribution of Data of Profile Curvature}
	\label{fig:profile-dis}
\end{figure}

PROFILE (Profile Curvature) is an important factor used in the study of Landslide Susceptibility Mapping. The curvature in the downslope direction along a line produced by the intersection of an imaginary vertical plane with the ground surface is known as Profile Curvature. The driving and resistive strains within a landslide in the direction of motion are affected by profile curvature. \citep{carson1972hillslope, Meten2015}. The data count distribution of the feature PROFILE is depicted in Figure \ref{fig:profile-dis}.

\subsection{PLAN (Plan Curvature)}

\begin{figure}[h!]
	\centering
	\includegraphics[scale = 0.6]{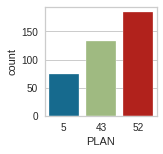}
	\caption{Count Distribution of Data of Plan Curvature (PLAN)}
	\label{fig:plan-dis}
\end{figure}

The curvature of topographic contours or the curvature of a line generated by the intersection of an imaginary horizontal plane with the ground surface is referred to as Plan Curvature (PLAN). The convergence or divergence of landslide material and water in the direction of landslide motion is controlled by plan curvature \citep{ohlmacher2007plan, Meten2015}.  The data count distribution of the feature PLAN is depicted in Figure \ref{fig:plan-dis}.

\subsection{CHANGE}

\begin{figure}[h!]
	\centering
	\includegraphics[scale = 0.6]{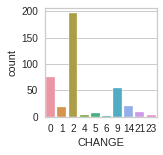}
	\caption{Count Distribution of Data of Change}
	\label{fig:change-dis}
\end{figure}

The loss of natural areas, notably forests, to urban or suburban development, or the loss of agricultural regions to urban or exurban development is referred to as land cover change (CHANGE) \citep{SEALEY201842}.
A number of studies have found that land cover change (CHANGE) is a notable and strongly influencing factor in the determination of the susceptibility of Landslides for a particular area \citep{PROMPER201411, restrepo2006landslides}. The data count distribution of the feature CHANGE is depicted in Figure \ref{fig:change-dis}.

\subsection{LANDUSE}

\begin{figure}[h!]
	\centering
	\includegraphics[scale = 0.6]{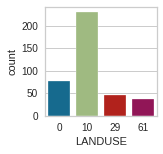}
	\caption{Count Distribution of Data of Landuse}
	\label{fig:landuse-dis}
\end{figure}

Land use is usually described as a sequence of human-performed activities on land with the goal of obtaining goods and advantages from the usage of land resources \citep{Reichenbach2014}. In hilly or mountainous areas, Land use change (LANDUSE) can increase or decrease the possibility of landslides with potential influence \citep{Chen2019}. The data count distribution of the feature LANDUSE is depicted in Figure \ref{fig:landuse-dis}.

\subsection{ELEVATION}

\begin{figure}[h!]
	\centering
	\includegraphics[scale = 0.6]{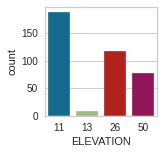}
	\caption{Count Distribution of Data of Elevation}
	\label{fig:elevation-dis}
\end{figure}

Landslide vulnerability is frequently assessed using elevation. The altitude of the terrain is referred to as elevation. According to \citet{dou2015optimization}, the ground at various heights will have varying levels of sensitivity which is a key factor to identify the probability of possible Landslide events.
The data count distribution of the feature ELEVATION is depicted in Figure \ref{fig:elevation-dis}.

\subsection{SLOPE}

\begin{figure}[h!]
	\centering
	\includegraphics[scale = 0.6]{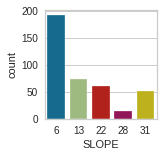}
	\caption{Count Distribution of Data of Slope Angle (SLOPE)}
	\label{fig:slope-dis}
\end{figure}

The angle measured between a horizontal plane and a particular location on the ground surface is known as the slope angle (SLOPE) \citep{WHITWORTH2011459}.
SLOPE is one of the most influential factors that can lead to causing serious landslides. SLOPE also has a notable correlation with other geological and topological factors which made it an early alarm to assess the susceptibility of landslides in the hilly parts of the world. In general, the likelihood of a landslide rises as the slope rises \citep{Meten2015}. The data count distribution of the feature SLOP is depicted in Figure \ref{fig:slope-dis}.

\subsection{ASPECT}
\begin{figure}[h!]
	\centering
	\includegraphics[scale = 0.6]{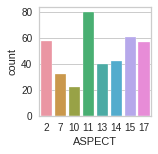}
	\caption{Count Distribution of Data of ASPECT}
	\label{fig:aspect-dis}
\end{figure}

The aspect at a location on the ground surface, according to some researchers, is the direction that the tangent plane passing through that point faces and is represented in degrees. The aspect, in its most basic form, is a data type that indicates the geographical direction in which the slopes grow \citep{tanoli2017spatial}. The data count distribution of the feature ASPECT is depicted in Figure \ref{fig:aspect-dis}.

\subsection{TWI}
\begin{figure}[h!]
	\centering
	\includegraphics[scale = 0.6]{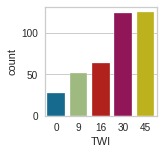}
	\caption{Count Distribution of Data of TWI}
	\label{fig:TWI-dis}
\end{figure}

The Topographic Wetness Index (TWI), also known as the compound topographic indicator, is a wetness index that measures steady-state conditions. It is widely used to quantify the influence of topography on hydrological processes which has a great influence on the occurrence of Landslides \citep{Mattivi2019}.
The data count distribution of the feature TWI is depicted in Figure \ref{fig:TWI-dis}.
\subsection{SPI}
\begin{figure}[h!]
	\centering
	\includegraphics[scale = 0.6]{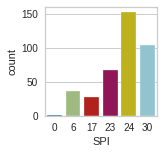}
	\caption{Count Distribution of Data of SPI}
	\label{fig:spi-dis}
\end{figure}

The erosive force of a stream or water flow is measured by the SPI (Stream Power Index). The slope and contributing area are used to calculate SPI. SPI approximates the locations on the landscape where gullies are more likely to form \citep{Abedin_Features}. The data count distribution of the feature SPI is depicted in Figure \ref{fig:spi-dis}.

\subsection{DRAINAGE}
\begin{figure}[h!]
	\centering
	\includegraphics[scale = 0.6]{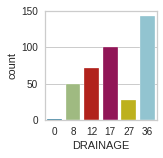}
	\caption{Count Distribution of Data of DRAINAGE}
	\label{fig:DRAINAGE-dis}
\end{figure}

DRAINAGE refers to distance to drainage network. It is usually noticed that area near to drainage network are more prone to landslides which makes it a very crucial feature for the study of landslides \citep{Abedin_Features}. The data count distribution of the feature DRAINAGE is depicted in Figure \ref{fig:DRAINAGE-dis}.
\subsection{NDVI}

\begin{figure}[h!]
	\centering
	\includegraphics[scale = 0.6]{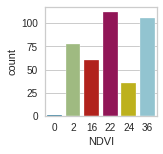}
	\caption{Count Distribution of Data of NDVI}
	\label{fig:NDVI-dis}
\end{figure}

The Normalized Difference Vegetation Index (NDVI) is used to calculate the density of green on a given plot of land. It measures vegetation by comparing the amount of near-infrared light reflected by vegetation to the amount of red light absorbed by vegetation. It has been identified as a good indicator of landslide susceptibility according to geotechnical researchers \citep{geosciences6040045}. The data count distribution of the feature NDVI is depicted in Figure \ref{fig:NDVI-dis}.

\subsection{RAINFALL}
\begin{figure}[h!]
	\centering
	\includegraphics[scale = 0.6]{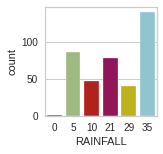}
	\caption{Count Distribution of Data of RAINFALL}
	\label{fig:RAINFALL-dis}
\end{figure}

The amount of rainfall in a particular hilly area is a great indicator of landslide susceptibility. Excessive rainfall is often considered as a potential trigger for sudden and destructive landslides in the hilly regions \citep{Abedin_Features}. Heavy rainfall can induce soil saturation, and debris flow can occur on certain slopes triggering the possibilty of rainfall induced landslides \citep{chen2017evaluating}. The data count distribution of the feature RAINFALL is depicted in Figure \ref{fig:RAINFALL-dis}.

\subsection{FAULTLINES}
\begin{figure}[h!]
	\centering
	\includegraphics[scale = 0.6]{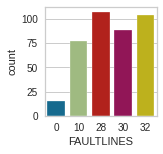}
	\caption{Count Distribution of Data of FAULTLINES}
	\label{fig:FAULTLINES-dis}
\end{figure}

Fault lines (FAULTLINES) are geological variables in a Lanslides research that suggest tectonic breaks and reduce rock strength. In general, areas closer to the Faultline are more prone to landslides than areas further away \citep{Regmi2014, Abedin_Features}. The data count distribution of the feature FAULTLINES is depicted in Figure \ref{fig:FAULTLINES-dis}.

\subsection{ROAD}
\begin{figure}[h!]
	\centering
	\includegraphics[scale = 0.6]{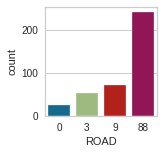}
	\caption{Count Distribution of Data of ROAD}
	\label{fig:ROAD-dis}
\end{figure}

ROAD refers to the distance from the road of land that is a crucial measure to assess the landslide susceptibility of an area. Roads assist to concentrate drainage, while road cuttings harm the slope structure. Landslides near roadways might occur if the required precautions are not taken \citep{chen2017evaluating}. The data count distribution of the feature ROAD is depicted in Figure \ref{fig:ROAD-dis}.

\subsection{GEOLOGY}
\begin{figure}[h!]
	\centering
	\includegraphics[scale = 0.6]{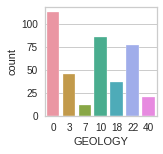}
	\caption{Count Distribution of Data of GEOLOGY}
	\label{fig:GEOLOGY-dis}
\end{figure}

Geology is concerned with the permeability and strength of a region's rocks and soil, and hence with landslides \citep{ayalew2005application}. Understanding the geology of land area is always considered to be a crucial factor for effective study of Landslides. The data count distribution of the feature GEOLOGY is depicted in Figure \ref{fig:GEOLOGY-dis}.

In this study, the abovementioned 15 features have been taken into consideration for building  a robust explainable machine learning model that best corroborates the prediction of Landslide Susceptibility adopting the geotechnical engineering domain. The study's features have also been analyzed in various state-of-the-art investigations, with a high level of correlation in terms of landslide susceptibility mapping.

\section{Experimental Results}
\label{sec:results}
In this section, we briefly explain the statistical and machine learning-based analysis employed on our dataset with the illustration of extensive experimental results to highlight our significant research findings and validate the proposal of an integrated automated framework for landslide susceptibility mapping.
\subsection{Normality Test}
The normality test is a statistical analysis-based method that is used to determine whether or not the distribution of data follows a Gaussian Distribution. In data mining, it is often important to comprehend the distribution of data that helps to discern whether to use parametric or non-parametric statistical methods for exploratory data analysis. If the data follows a Gaussian distribution, then parametric statistical methods are incorporated, else non-parametric statistical methods are incorporated \citep{aitchison1982statistical}. In this context, to understand the data distribution of our feature variables involved in Landslide Susceptibility prediction we employed a popular and reliable normality evaluation method \citep{compare_normality}, Sharpio-Wilks, that generates p-values based on test statistics (W-statistic). The W-statistic or test statistic is computed as \citep{equation_normality};
\begin{equation}\label{noramility}
W={\left(\sum _{i=1}^{n}a_{i}x_{(i)}\right)^{2} \over \sum _{i=1}^{n}(x_{i}-{\overline {x}})^{2}}
\end{equation}
Where $W$ denotes the test statistic value, $x_{{(i)}}$ denotes the $ith$ order statistic and $\overline {x}$  denotes the mean of the sample. And also, $a_{i}$, the coefficient can be stated as;
\begin{equation}\label{norm_1}
(a_{1},\dots ,a_{n})={m^{\mathsf {T}}V^{-1} \over C}
\end{equation}
In equation \ref{norm_1}, $a_{i}$ coefficients not only depend on the sample size $m$ but also the same for all samples of that size. Even it can be calculated beforehand and indexed in a tabular format.
Here $C$ is a vector norm that can be donated by;
\begin{equation}\label{norm_2}
C=\|V^{-1}m\|=(m^{\mathsf {T}}V^{-1}V^{-1}m)^{1/2}
\end{equation}

 Also, $m$ represents the sample size and $V$ represents the expected value $i-th$ order statistic under normality in Equations \ref{norm_1} and \ref{norm_2}. The following Table \ref{tab:sharpio-wilk} represents the W-Statistic values computed through the Sharpio-Wilks test for each individual feature variable of our dataset. 

\begin{table}[h]
\centering
\begin{tabular}{|r|r|r|r|}
\hline
\textbf{Feature} & \textbf{W-Stats} & \textbf{Feature} & \textbf{W-Stats} \\ \hline
\textbf{PROFILE}    & 0.752 & \textbf{SLOPE}    & 0.771 \\ \hline
\textbf{PLAN}       & 0.636 & \textbf{ASPECT}   & 0.864 \\ \hline
\textbf{CHANGE}     & 0.685 & \textbf{TWI}      & 0.868 \\ \hline
\textbf{LANDUSE}    & 0.665 & \textbf{SPI}      & 0.749 \\ \hline
\textbf{ELEVATION}  & 0.735 & \textbf{DRAINAGE} & 0.809 \\ \hline
\textbf{NDVI}       & 0.857 & \textbf{ROAD}     & 0.642 \\ \hline
\textbf{RAINFALL}   & 0.818 & \textbf{GEOLOGY}  & 0.847 \\ \hline
\textbf{FAULTLINES} & 0.684 &                   &       \\ \hline
\end{tabular}
\caption{Sharpio-Wilk Test Statistic of Feature Variables}
\label{tab:sharpio-wilk}
\end{table}

To apply the Sharpio-Wilk test to our feature variable, we formulate a null hypothesis a; the population from which the sample was drawn follows a normal distribution and the alternative hypothesis as; the population from which the sample was drawn does not follow a normal distribution. To calculate W-Statistic and understand the normality, we incorporated the Scipy tool \citep{virtanen2020scipy} using their "sharpio" function. The tool returned the W-statistics scores and P-values scores for each feature variable. P-values illustrate from a given data sample how likely the data was drawn from a Gaussian distribution based on a certain threshold. Evidently, a p-value less than or equal to the threshold value of 0.05 delineates that the data sample does not follow a normal or Gaussian distribution and a p-value greater than 0.05 delineates that data was drawn from a normal distribution. In this study, based on the normality test analysis, we found that the data samples from any individual independent variable were not drawn from a normal distribution as the p-value for each feature was so much less than 0.05 making strong ground for rejecting the null hypothesis. However, numerous characteristics of the dataset are continuous in nature when it comes to the area of Geotechnical Engineering, which is poorly reflected by the data distribution. According to this research, a linear machine learning model would not be the best match to capture complicated multi-co-linearity difficulties when applied to a different dataset's global point of view. Considering the analysis, we successfully applied the geotechnical engineering domain to identify the best potential machine learning solution throughout our study.
\subsection{Chi-Square Test}
In data mining and inference-based research, it is important to understand whether or not a certain feature contributes to the final outcome. The statistical feature significance test helps us to identify and select features that strongly corroborate the final prediction. In Geo-technical Science, identifying key factors that best predict Landslide Susceptibility is an indispensable part. In our study, we decided to employ a non-parametric statistical significance-based test, the Chi-Square test, to identify the importance of our feature variables in Landslide Susceptibility prediction based on the normality test results. The Chi-Square test is a non-parametric statistical significance analysis method that is suitable for analyzing the significance of independent variables. To state mathematically \citep{Chi-square, Chi-square-2},

\begin{equation}\label{chi-equation}
    {X_c}^2 = \sum_{i=1}^{N} {\frac{(O_i -E_i)^2}{E_i}}
\end{equation}

Here, $c$ denotes "Degree of Freedom", $O$ denotes "observed value" and $E$ denotes "expected value", and $i$ is the ”ith” position is in the contingency table. The value of $c$ is computed through,

\begin{equation}\label{degfreedom}
	c = (N_d - 1) \times (N_f - 1)
\end{equation}

Here, $N_d$ denotes \textit{"Number of Data Instances} and $N_f$ denotes \textit{"Number of Features"}.
Afterward, we determine the statistically significant P-values for the independent variable against the dependent variable using the computed Chi-Square and degree of freedom values. The p-values computed through the Chi-square test are shown in Table \ref{tab:chi-square}.

According to the p-value threshold of 0.05, we can infer that all of our feature variables show statistical evidence to espouse the prediction of Landslide Susceptibility. 
\begin{table}[H]
\centering
\begin{tabular}{|r|r|r|r|}
\hline
\textbf{Feature} & \textbf{P-Value} & \textbf{Feature} & \textbf{P-Value} \\ \hline
\textbf{PROFILE}      & 3.19E-94         & \textbf{SLOPE}        & 9.73E-291        \\ \hline
\textbf{PLAN}         & 1.13E-123        & \textbf{ASPECT}       & 4.04E-17         \\ \hline
\textbf{CHANGE}       & 1.24E-57         & \textbf{TWI}          & 2.89E-308        \\ \hline
\textbf{LANDUSE}      & 6.56E-158        & \textbf{SPI}          & 8.93E-11         \\ \hline
\textbf{ELEVATION}    & 7.11E-162        & \textbf{DRAINAGE}     & 7.77E-17         \\ \hline
\textbf{NDVI}         & 9.96E-98         & \textbf{ROAD}         & 0                \\ \hline
\textbf{RAINFALL}     & 6.00E-82         & \textbf{GEOLOGY}      & 4.32E-88         \\ \hline
\textbf{FAULTLINES}   & 1.27E-48         &                       &                  \\ \hline
\end{tabular}
\caption{p-value Scores From Chi-Square Test}
\label{tab:chi-square}
\end{table}

Here, a p-value of less than or equal to 0.05 indicates that the particular categorical feature variable significantly contributes to the classification of Landslide Susceptibility with strong statistical evidence.

\subsection{Evaluation Metrics}
In our study, to evaluate the performance of our machine learning models, we considered the analysis of below mentioned popularly used evaluation metrics.
\begin{itemize}
	\item True Positive (TP): The case when the certain area is Landslide Susceptible and the model also classified as Landslide Susceptible. 
	\item False Positive (FP): The case when the certain area is not Landslide Susceptible but the model classified as Landslide Susceptible. 
	\item True Negative (TN): The case when the certain area is Not Landslide Susceptible and the model also classified as Not Landslide Susceptible. 
	\item False Negative (FN): The case when the certain area is Landslide Susceptible but the model classified as Not Landslide Susceptible.
	\item Accuracy: It defines correctly classified areas with susceptible to Landslides. It can be computed as;
	\begin{equation}\label{accuracy-eq}
		Accuracy = \frac{TP + TN}{TP + FP + TN + FN}
	\end{equation}
	
	\item Recall: It is defined as the ratio of the number of positive samples that have been correctly predicted as Landslide Susceptible corresponding to all Landslide Susceptible samples in the data. It can computed as;
	\begin{equation}
    Recall = \frac{TP}{TP + FN}
    \end{equation}
    \item Precision: It is defined as the ratio of the number of positive samples that have been correctly predicted as Landslide Susceptible corresponding to all samples predicted as Landslide Susceptible. It can be computed as;
    \begin{equation}
    Precision = \frac{TP}{TP + FP}
    \end{equation}
    \item F1-Score: It is delineated as the term that balances between recall and precision. It can be defined as;
    \begin{equation}
    F1-Score = 2 \times \frac{Recall \times Precision}{Recall + Precision}
    \end{equation}
	
\end{itemize}

\subsection{Evaluation Stage 1: Optimization of Algorithms}
\label{sec:ev}
In this section, we have optimized the performance of our Machine Learning classifiers using the Grid Search method to find the best combination of Hyper-parameters for individual classifiers with 10-Fold Stratified Cross Validation to reduce the overfitting issue. We have employed the Scikit-Learn implementation in Python for SVM, KNN, Adaboost and Logistic Regression. And the Python Library of XgBoost for implementing the Extreme Gradient Boosting Algorithm. Grid Searching was performed by incorporating the GridSearchCV method from the popular Scikit-Learn library. Grid Search is an exhaustive searching method popularly used for hyperparameter tuning of Machine Learning algorithms. This method incorporates grid-based parameter search by which it computes every possible combination of parameters from a given set of values. It helps to optimize the performance along with reducing the overfitting issue of Machine Learning algorithms to build an efficient classifier based on certain data. In our study, to optimize the machine learning algorithms for landslide susceptibility prediction, we have incorporated the Grid Search method for finding the best hyperparameters for every individual classifier validating by 10-Fold Stratified Cross-Validation of Weighted F1 Scores. The grid of hyperparameters which has been optimized with Grid Search depicted in Table \ref{tab:grid}.

\begin{table*}[h!]
\centering
\begin{tabular}{|l|l|l|l|}
\hline
\textbf{Classifier} &
  \textbf{HP} &
  \textbf{Definition} &
  \textbf{Parameters Grid} \\ \hline
XgBoost &
  max\_depth &
  Maximum Depth of a Tree &
  {[}2,3,5,6,8{]} \\ \cline{2-4} 
 &
  n\_estimators &
  Number of trees &
  \begin{tabular}[c]{@{}l@{}}{[}500, 1500, \\ 3000, 5000{]}\end{tabular} \\ \cline{2-4} 
 &
  learning\_rate &
  Learning Rate &
  \begin{tabular}[c]{@{}l@{}}{[}0.01, 0.1, \\ 0.05, 0.3, 0.5{]}\end{tabular} \\ \cline{2-4} 
 &
  gamma &
  Regularization Parameter &
  \begin{tabular}[c]{@{}l@{}}{[}0, 0.1, 0.5, \\ 1, 2{]}\end{tabular} \\ \cline{2-4} 
 &
  subample &
  Percentage of Training Rows &
  \begin{tabular}[c]{@{}l@{}}{[}0.5, 0.7, \\ 0.8, 0.9, 1{]}\end{tabular} \\ \hline
KNN &
  n\_neighbors &
  Number of Neighbours &
  \begin{tabular}[c]{@{}l@{}}{[}3, 5, 7, 9, \\ 11, 13{]}\end{tabular} \\ \cline{2-4} 
 &
  p &
  \begin{tabular}[c]{@{}l@{}}Exponent of\\ Minkowski distance\end{tabular} &
  {[}1, 2{]} \\ \hline
LR       & C             & Regularization Parameter & \begin{tabular}[c]{@{}l@{}}{[}0.001, 0.01, 0.1, \\ 1, 10, 100{]}\end{tabular}             \\ \cline{2-4} 
 &
  solver &
  Algorithm to use &
  \begin{tabular}[c]{@{}l@{}}{[}'newton-cg', 'lbfgs', \\ 'liblinear', 'sag', 'saga'{]}\end{tabular} \\ \cline{2-4} 
 &
  penalty &
  Regularization Algorithm &
  {[}'l1', 'l2', 'elasticnet'{]} \\ \hline
SVM &
  C &
  Regularization Parameter &
  {[}0.01, 0.1, 1, 10, 100{]} \\ \cline{2-4} 
 &
  kernel &
  Kernel Trick &
  {[}'linear', 'rbf'{]} \\ \hline
Adaboost & n\_estimators & Number of trees          & \begin{tabular}[c]{@{}l@{}}{[}0.001, 0.01, 0.1, \\ 0.15, 0.2, 0.3, 0.5, 1{]}\end{tabular} \\ \cline{2-4} 
 &
  learning\_rate &
  Learning Rate &
  \begin{tabular}[c]{@{}l@{}}{[}10, 50, 100, \\ 500, 1000, 1500, 3000{]}\end{tabular} \\ \hline
\end{tabular}
\caption{Range of Hyperparameters (HP) and Definition}
\label{tab:grid}
\end{table*}

In this stage of evaluation, all 15 feature variables have been employed for training and testing the individual classifiers. The best combinations of hyper-parameters of the machine learning classifiers obtained through Grid Searching are illustrated in Table ~ \ref{tab:best-hyper}.

\begin{figure}[h!]
	\centering
	\includegraphics[scale=0.4]{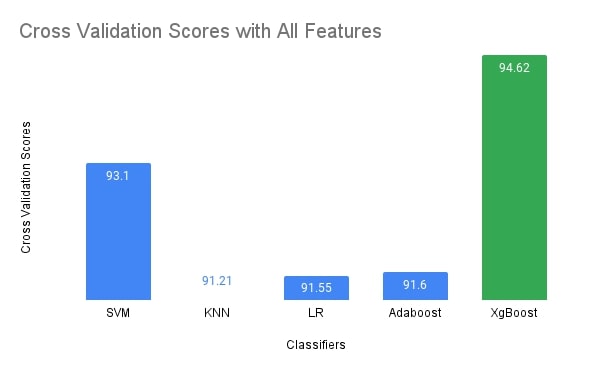}
	\caption{Cross Validation Scores With All Features}
	\label{cv-all}
\end{figure}

\begin{table}[h!]
\centering
\begin{tabular}{|r |r |}
\hline
\textbf{Classifier} & \textbf{Hyperparameters}                                                      \\ \hline
XgBoost              & \begin{tabular}[c]{@{}r@{}}max\_depth=3\\ n\_estimators=3000\\ learning\_rate=0.1\\ gamma=0\\ subample=0.7\end{tabular} \\ \hline
KNN                 & \begin{tabular}[c]{@{}r@{}}n\_neighbors =7\\ p = 1\end{tabular}               \\ \hline
Logistic  Regression & \begin{tabular}[c]{@{}r@{}}C=0.01\\ solver='l2'\\ penalty='newton-cg'\end{tabular}                                      \\ \hline
SVM                 & \begin{tabular}[c]{@{}r@{}}C=10\\ kernel=rbf\end{tabular}                     \\ \hline
Adaboost            & \begin{tabular}[c]{@{}r@{}}n\_esitmators=1000\\ learning\_rate=1\end{tabular} \\ \hline
\end{tabular}
\caption{Best Combination of Hyperparameters}
\label{tab:best-hyper}
\end{table}
A graphical log-scale comparison of 10-Fold Stratified Cross-validation Scores with the obtained best hyper-parameters combination of the machine learning classifiers employed in our study is represented in the Figure ~\ref{cv-all}.

 After selecting the best set of hyper-parameters using a 10-Fold Stratified Cross Validation, we sampled our dataset into a training and testing ratio of 67:33 to perform additional hold validation for individual classifiers. The Confusion Matrix results of the machine learning classifiers on the test dataset from the hold-out validation set are presented in Table \ref{tab:cm-all}.

\begin{table}[h!]
\centering
\begin{tabular}{|c|c|c|c|c|c|}
\hline
                        & \textbf{SVM} & \textbf{KNN} & \textbf{LR} & \textbf{AdaBoost} & \textbf{XgBoost} \\ \hline
\textbf{TP}  & 57           & 60           & 58          & 57                & 59               \\ \hline
\textbf{FP} & 8            & 8            & 3           & 3                 & 5                \\ \hline
\textbf{TN}  & 57           & 57           & 62          & 60                & 60               \\ \hline
\textbf{FN} & 8            & 5            & 7           & 8                 & 6                \\ \hline
\end{tabular}
\caption{Confusion Matrix Before Feature Reduction}
\label{tab:cm-all}
\end{table}

Precision, Recall, F1-Score and Accuracy scores of the machine learning classifiers is presented on the Table ~\ref{tab:ev-all}.

\begin{table}[h!]
\centering
\begin{tabular}{|c|c|c|c|c|c|}
\hline
\textbf{Metrics} & \textbf{SVM} & \textbf{KNN} & \textbf{LR} & \textbf{AdaB} & \textbf{XgB} \\ \hline
\textit{Accuracy}      & 87.69 & 90    & 92.31 & 90    & 91.54 \\ \hline
\textit{Precision (N)} & 87.69 & 91.94 & 89.86 & 88.24 & 90.91 \\ \hline
\textit{Recall (N)}    & 87.69 & 87.69 & 95.38 & 92.31 & 92.31 \\ \hline
\textit{F1 Score (N)}  & 87.69 & 89.76 & 92.54 & 90.23 & 91.6  \\ \hline
\textit{Precision (P)} & 87.69 & 88.24 & 95.08 & 91.94 & 92.19 \\ \hline
\textit{Recall (P)}    & 87.69 & 92.31 & 89.23 & 87.69 & 90.77 \\ \hline
\textit{F1 Score (P)}  & 87.69 & 90.23 & 92.06 & 89.76 & 91.47 \\ \hline
\end{tabular}
\caption{Performance Evaluation with All Features with Best Hyper-parameters of Classifiers}
\label{tab:ev-all}
\end{table}

In this context, based on the comparison of the performances of the individual machine learning classifiers, it is clear that the optimized version of XgBoost(Extreme Gradient Boosting) is outperforming all other Machine Learning classifiers in evaluation criteria based on 10 Fold Stratified Cross Validation Mean F1 Weighted Scores. However, the findings on the test set demonstrate that the Logistic Regression model outperforms all other classifiers. But, as the logistic regression model does not outperform in terms of 10 Fold Cross Validation Scores, it may be deduced that the test dataset scores do not completely reflect the actual efficiency due to the lack of the bulk of data samples. Apart from that, based on F1 Scores on test data samples, XgBoost is still showing the second-best performance. Moreover, tree-based gradient-boosted methods have better explainability than other machine-learning classifiers. Thus, in this evaluation stage, Optimized Extreme Gradient Boosting is identified as the most efficient algorithm to classify Landslide Susceptible areas using the 15 feature variables from the utilized dataset.

\subsection{Evaluation Stage 2: TreeSHAP Analysis}
\label{sec:ev2}
As optimized XgBoost is outperforming all other machine learning classifiers besides it has robust explainability, we further decided to proceed with the optimized XgBoost model for further analysis. In this stage of evaluation, we have integrated the TreeSHAP method with the optimized XgBoost (Extreme Gradient Boosting) Classifier(our best-performing classifier from the previous evaluation stage) to understand the prediction criteria of XgBoost and the level of contribution of individual features that strongly corroborate in Landslide Susceptibility prediction. Explainable Artificial Intelligence is an essential demand of the present time as the complex and deep architectures of these machine learning models make it hard to interpret by engendering the BlackBox problem which is defined as not being able to detect where the model is actually looking or how much the individual features are contributing for a certain prediction.  Considering the exigency and sensitivity involved in the study of Landslide Susceptibility, we interpreted the performance of XgBoost by analyzing the SHAP Feature Importance and SHAP Summary Plot.

\subsubsection{SHAP Feature Importance}
\begin{figure}[h!]
	\centering
	\includegraphics[scale = 0.4]{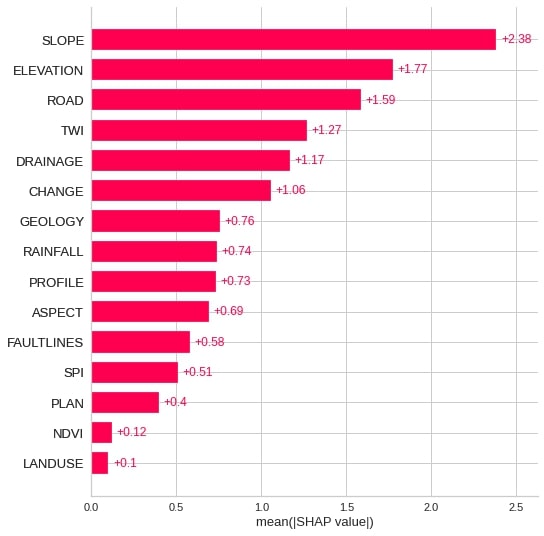}
	\caption{SHAP Feature Importance}
	\label{SHAP_Feature_Importance}
\end{figure}

The SHAP Feature Importance plot illustrates the mean absolute Shapley values for individual features. Figure ~\ref{SHAP_Feature_Importance} presents a graphical illustration of the mean absolute Shapley values of all the features used for training the optimized Extreme Gradient Boosting (XgBoost) classifier before predicting Landslide Susceptibility. It is a bar plot where features are sorted in descending order based on mean absolute SHAP values. Features with large absolute values like SLOPE, ELEVATION, ROAD, and TWI have a significant impact on the prediction to support the optimized XgBoost model for detecting Landslide Susceptible areas efficiently. On the other hand, FAULTLINES, PLAN, SPI, NDVI, and LANDUSE have low absolute mean values indicating a very low influence on the model's performance.

\subsubsection{SHAP Summary Plot}
\begin{figure}[h!]
	\centering
	\includegraphics[scale = 0.4]{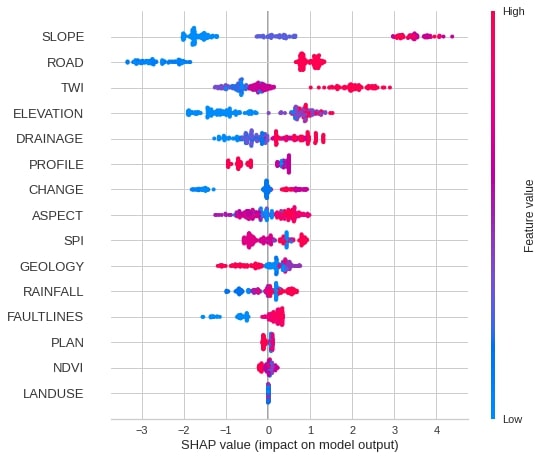}
	\caption{SHAP Summary Plot}
	\label{SHAP_Summary_Plot}
\end{figure}

The SHAP Summary Plot illustrated in Figure ~\ref{SHAP_Summary_Plot} combines feature importance with feature effects for the optimized XgBoost model of Landslide Susceptibility prediction.
In the following summary plot, each point is a Shapley value for a feature and a data instance. Here, in the graph, the vertical axis delineates features sorted according to the level of influence in the model's prediction and the horizontal axis delineates Shapley values indicating the positive or negative correlation of the feature of an instance with the target variable. To understand the values of features graphically, a color-based comparison is drawn. The overlapping dots are jittered in the vertical axis direction which gives us an idea of the Shapley value distribution per feature variable. The features are ranked in order of significance. The following summary plot strongly illustrates; SLOPE, ROAD, and TWI as the most significant features with widespread distribution along the horizontal axis supporting
efficient prediction for Landslide Susceptibility. In contrast, instances of PLAN, NDVI and LANDUSE, and some other features are not  well distributed over the horizontal axis and possess a Shapley value of near 0, indicating very less or almost no influence on the model's performance for Landslide Susceptibility Prediction.

The study of features' contributions to the assessment of Landslide Susceptible regions derived from TreeSHAP analysis can be validated by considering evidence from state-of-the-art studies by geoscience researchers. 
Firstly, \citet{ccellek2020effect} has emphasized the importance of measuring Slope Angle (SLOPE) for classifying Landslides with strong evidence. Consequently, the optimized XgBoost model proposed in our study is also utilizing the SLOPE feature as the top most significant feature for the prediction of Landslides. Furthermore, ELEVATION has a great correlation with SLOPE in the study of Landslides and helps researchers to understand the vulnerability of an area towards Landslides \citep{Rabby-Dataset}.  This fact has also been reflected in our investigation of Landslides through TreeSHAP analysis. Additionally, TWI and ROAD are also top contributing factors for assessing the susceptibility of Landslides which have been reflected in our model-building process and studies by several geotechnical researchers \citep{chen2017evaluating, Abedin_Features, Rabby-Dataset}. Thus, the following analysis strengthens the hypothesis of this study and supports the proper adoption of the Geoscience and Geotechnical Engineering domain at every stage of our proposed research framework.

\subsection{Evaluation Stage 3: Improved Landslide Feature Selection}
\begin{figure*}[h!]
	\centering
	\includegraphics[scale = 0.4]{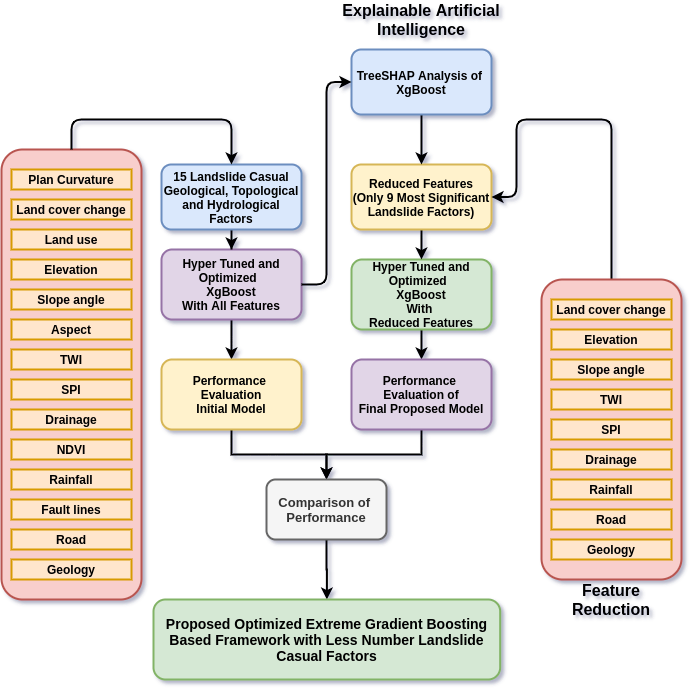}
	\caption{The Complete Research Workflow for Building the Proposed Integrated XgBoost Model for Landslide Susceptibility Mapping using Less Number Landslide Causal Factors}
	\label{xgboost-workflow}
\end{figure*}

In machine learning-based solutions, feature reduction is extremely important. An efficient model that can forecast Landslides with high accuracy using only a few characteristics is a gift in the domain of geotechnical engineering for Landslide Susceptibility Mapping.
From the evaluation scores of previous stages, the optimized version of XgBoost seems to be the best fit for portending landslide susceptibility. So, in this stage of the experimental setup, we have elected XgBoost as our proposed model for integration in landslide susceptibility mapping and performed feature reduction for XgBoost using the analysis of SHAP values. For this, we have retrained XgBoost with Grid Searching and eliminated 6 features including ASPECT, FAULTLINES, SPI, PLAN, NDVI and LANDUSE which has comparatively a very low influence in the prediction of XgBoost model's performance according to TreeSHAP analysis.  The overfall research methodology of the proposed integrated optimized XgBoost model which can efficiently predict landslide susceptibility by using less number of landslide causal factors is depicted in Figure \ref{xgboost-workflow}.


\begin{figure}[h!]
	\centering
	\includegraphics[scale=0.6]{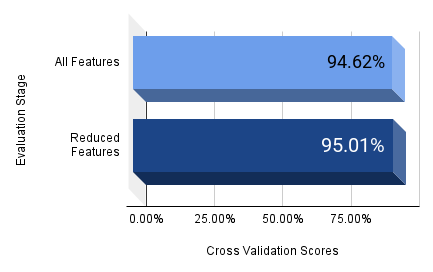}
	\caption{Comparison of Cross-Validation Scores of XgBoost with and without feature reduction}
	\label{cv-comp}
\end{figure}

Comparison of 10 Fold Stratified Cross Validation Mean Weighted F1 Scores before and after reduction of features illustrated in Figure~\ref{cv-comp}.

It is clearly visible that after feature reduction the Cross Validation score of optimized XgBoost has been increased from 94.62\% to 95.01\%. 

\begin{table}[h!]
\centering
\begin{tabular}{|c|c|}
\hline
\textbf{Parameter Name} & \textbf{Value} \\ \hline
gamma          & 0     \\ \hline
learning\_rate & 0.1   \\ \hline
max\_depth     & 3     \\ \hline
n\_estimators  & 1500  \\ \hline
subsample      & 1     \\ \hline
\end{tabular}
\caption{Final Set of Hyperparameters of proposed optimized XgBoost model with Reduced Features}
\label{tab:final-hyper}
\end{table}

Table \ref{tab:final-hyper} delineates the final hyperparameter settings of the XgBoost model with reduced features.

\begin{figure}[h!]
	\centering
	\includegraphics[scale=0.4]{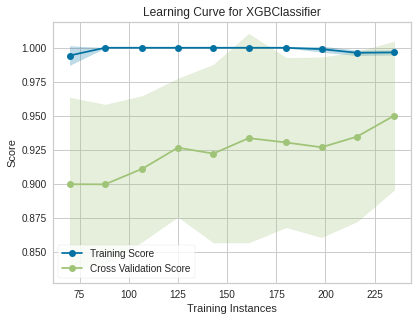}
	\caption{Learning Curve of XgBoost}
	\label{learning-curve-xgb}
\end{figure}

Learning Curve helps us to determine whether or not increasing the amount of data is going to ameliorate the performance of the model. The learning rate curve of the XgBoost classifier from Figure ~\ref{learning-curve-xgb} clearly depicts that there is room for improvement in the performance with more data in the case of the optimized XgBoost model with reduced features. The point where the two lines in the graphs will converge, after that particular point the room for improvement with more data will be narrowed down. 

\begin{figure}[h!]
	\centering
	\includegraphics[scale=0.4]{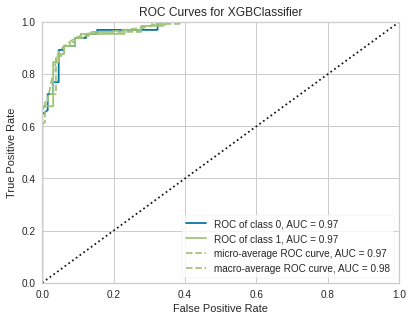}
	\caption{ROC Curve of XgBoost}
	\label{roc-xgb}
\end{figure}



\section{Discussion}
\label{sec:discussion}
In this study, after a rigorous analysis of machine learning classifiers for predicting Landslides in an automated manner, it is found that the optimized version of the XgBoost classifier is a promising method that successfully adopts geotechnical engineering domain based on the outperforming results of Table \ref{tab:ev-all} and Figure \ref{cv-all}. Considering the massive potential of the XgBoost model, this study further extended the research to explain the prediction criterion of the XgBoost model by integrating TreeSHAP analysis in the motivation of solving the blackbox problem with Explainable Artificial Intelligence. According to the TreeSHAP analysis, 6 features were eliminated from the feature set, and the XgBoost model was retrained along with exhaustive hyperparameter tuning. Here, after feature elimination, the 10 Fold Cross Validation Weighted F1 Score of the XgBoost model has increased to 95.01\%. It corroborates that feature reduction has been beneficial for the XgBoost model. Moreover, the optimized XgBoost model can generate outperforming results with even fewer features. 

Reduction of features is a crucial contribution presented in this study that opens doors for geoscience researchers to conduct effective studies on Landslides utilizing fewer features. The features that have been eliminated in the final model-building process in this study can reduce the cost of industry-level works on the domain of Landslide Susceptibility prediction. For example, to compute the features like LANDUSE and NDVI, high-tech remote and satellite-based technologies are required which increases the cost of the investigation and demands more time and technological manpower. In the primary stage, if the susceptibility of Landslide can be predicted without the need of these features which have high costs for computation purposes it is going to be a great help for geotechnical researchers for both industry and academia. The optimized version of Xgboost with the proposed architecture is outperforming other classifiers and previous settings of the model even after eliminating the above-mentioned 6 features. Due to the elimination of these 6 features, it would be possible to save more cost and time along with better utilization of the existing technological resources involved in the Landslide study.


Comparatively, optimized Extreme Gradient Boosting has outperformed all other machine learning classifiers even with a reduction of 40\% of the features with best hyper-parameters with 10 Fold Stratified Cross-Validation Weighted F1 Scores of 95.01\% and ROC-AUC score of 97\%. It has also helped us to identify eloquent features that strongly corroborate the prediction of Landslide Susceptibility. In this context, we prefer to adopt the optimized version of Extreme Gradient Boosting (XgBoost) with the identified eloquent features for Landslide Susceptibility Prediction using data-driven analysis. 

\begin{figure}[h!]
	\centering
	\includegraphics[scale=0.4]{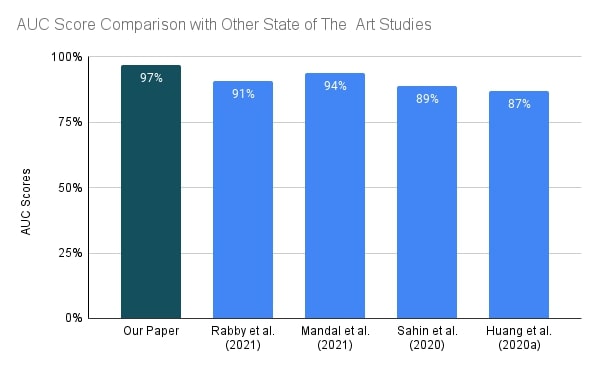}
	\caption{AUC Score Comparison with Other State of Art Studies}
	\label{auc-comparison}
\end{figure}
From Figure \ref{auc-comparison}, in comparison with the recent state-of-the-art studies for Landslide Susceptibility Prediction using Artificial Intelligence based methods, the optimized XgBoost model with Reduced Features outlined in our study, has outperformed the best model from the study of \citet{Rabby-Dataset} by almost 6\% , \citet{kanu-geo} by almost 4\%, \citet{Sahin-AI-tool} by almost 8\% and \citet{HUANG-ML2020} by almost 10\% in terms of AUC scores. Here, we utilized the exact same dataset as \citet{Rabby-Dataset} for our experimental evaluations. \citet{Sahin-AI-tool}, \citet{kanu-geo} and \citet{HUANG-ML2020} did not use the exactly same dataset as us, but their goal and scope of the study highly coincide with our research directions, in terms of methodologies and evaluations.
\section{Conclusion}
\label{sec:conclusion}
Landslides are another name of nightmare for people living in hilly areas from all over the world which causes a huge number of causalities every year exacerbating socio-economic conditions along with unwanted death tolls. An early prediction of Landslide Susceptibility can be a great blessing for mankind. Realizing this exigency, we have performed extensive analysis to predict Landslide Susceptible areas with state-of-the-art Artificial Intelligence based methods. In our study, we have successfully identified the eloquent features which are most useful for an automatic and early prediction of Landslide Susceptibility with Machine Learning algorithms, through the integration of Explainable and Interpretable Artificial Intelligence, a milestone achieved in the study of Landslides and Geo-Technical science. Besides, due to the optimization of models and even with the reduction of features our best-performing model has outperformed the recent similar state of art studies and the previous study with similar datasets also. Moreover, this study also highlights that machine learning-based feature selection is more suitable than the statistical analysis-based feature selection in the domain of Landslide study. The proposed model of this study is a cost-effective solution for the assessment and early prediction of Landslides at the primary investigation with fewer geological and topological features.  The reduction of geological, topological, and hydrological features would allow geotechnical researchers to conduct more economical and efficient research in less time. In the future, we would like to collect more empirical data on Landslides and integrate state of art deep architectures including Generative Adversarial Networks (GANs) for more efficient and early prediction of Landslides to address the problem of Landslide Susceptibility Mapping for saving mankind from sudden catastrophes.

\section*{Declarations}
\subsection*{Funding}
The authors did not receive support from any organization for the submitted work.
\subsection*{Conflicts of interest/Competing interests}
All authors certify that they have no affiliations with or involvement in any organization or entity with any financial interest or non-financial interest in the subject matter or materials discussed in this manuscript.


\section*{Acknowledgment}
We would like to extend our heartfelt gratitude to
Dr. Fahim Irfan Alam, Ph.D., an Associate Professor at the University
of Chittagong and a Post-Doctoral Research Fellow at the University
of New South Wales (UNSW), Sydney, Australia. Dr. Fahim, an expert
and experienced researcher in the domains of Machine Learning, Neural Networks, and Computer Vision, deserves our utmost appreciation
for his invaluable suggestions, meaningful discussions, and necessary
intellectual reviews, which have significantly contributed to elevating
the quality and performance of this research study

\end{document}